\documentclass[10pt,twocolumn,letterpaper]{article}

\usepackage[pagenumbers]{cvpr} 

\usepackage{url}
\usepackage{graphicx} 
\usepackage{booktabs}
\usepackage{makecell}

\usepackage{array}
\usepackage{multirow}
\usepackage{colortbl}

\usepackage[dvipsnames]{xcolor}

\usepackage[ruled]{algorithm2e}
\usepackage{bbm}
\usepackage{bm}
\usepackage{color}

\usepackage{diagbox}
\usepackage{bbding}

\usepackage{tabularx}
\usepackage{dashrule}
\usepackage[accsupp]{axessibility} 

\definecolor{cvprblue}{rgb}{0.21,0.49,0.74}
\usepackage[pagebackref,breaklinks,colorlinks,allcolors=cvprblue]{hyperref}

\def\paperID{1438} 
\def\confName{CVPR}
\def\confYear{2025}

\title{PyramidDrop: Accelerating Your Large Vision-Language Models via\\ Pyramid Visual Redundancy Reduction}

\author{
Long Xing$^{1,2}$, Qidong Huang$^{1,2}$, Xiaoyi Dong$^{2,3}$, Jiajie Lu$^{1,2}$, Pan Zhang$^{2}$, Yuhang Zang$^{2}$\\ Yuhang Cao$^{2}$, Conghui He$^{2}$, Jiaqi Wang$^{2}$, Feng Wu$^{1}$, Dahua Lin$^{2}$\\
$^1$University of Science and Technology of China \quad 
$^2$Shanghai AI Laboratory \quad
$^3$CUHK
}

\begin{document}
\maketitle
\begin{abstract}
In large vision-language models (LVLMs), images serve as inputs that carry a wealth of information. As the idiom ``A picture is worth a thousand words" implies, representing a single image in current LVLMs can require hundreds or even thousands of tokens. This results in significant computational costs, which grow quadratically as input image resolution increases, thereby severely impacting the efficiency. Previous approaches have attempted to reduce the number of image tokens either before or within the early layers of LVLMs. However, these strategies inevitably result in the loss of crucial image information. To address this challenge, we conduct an empirical study revealing that all visual tokens are necessary for LVLMs in the shallow layers, and token redundancy progressively increases in the deeper layers.
To this end, we propose PyramidDrop, a visual redundancy reduction strategy for LVLMs to boost their efficiency in both inference and training with neglectable performance loss. Specifically, we partition the LVLM into several stages and drop part of the image tokens at the end of each stage with a pre-defined ratio. The dropping is based on a lightweight similarity calculation with a negligible time overhead. Extensive experiments demonstrate that PyramidDrop can achieve over 40\% training time reduction and 55\% inference FLOPs acceleration on leading LVLMs like LLaVA-NeXT, maintaining comparable multi-modal performance. Besides, PyramidDrop can also serve as a plug-and-play strategy to accelerate inference in a free way, with better performance and lower inference cost than counterparts. This project is available at \url{https://github.com/Cooperx521/PyramidDrop} to serve as a pivotal resource for advancing the community. 

\end{abstract}

\begin{figure*}[t]
\centering
\includegraphics[width=1.0\linewidth]{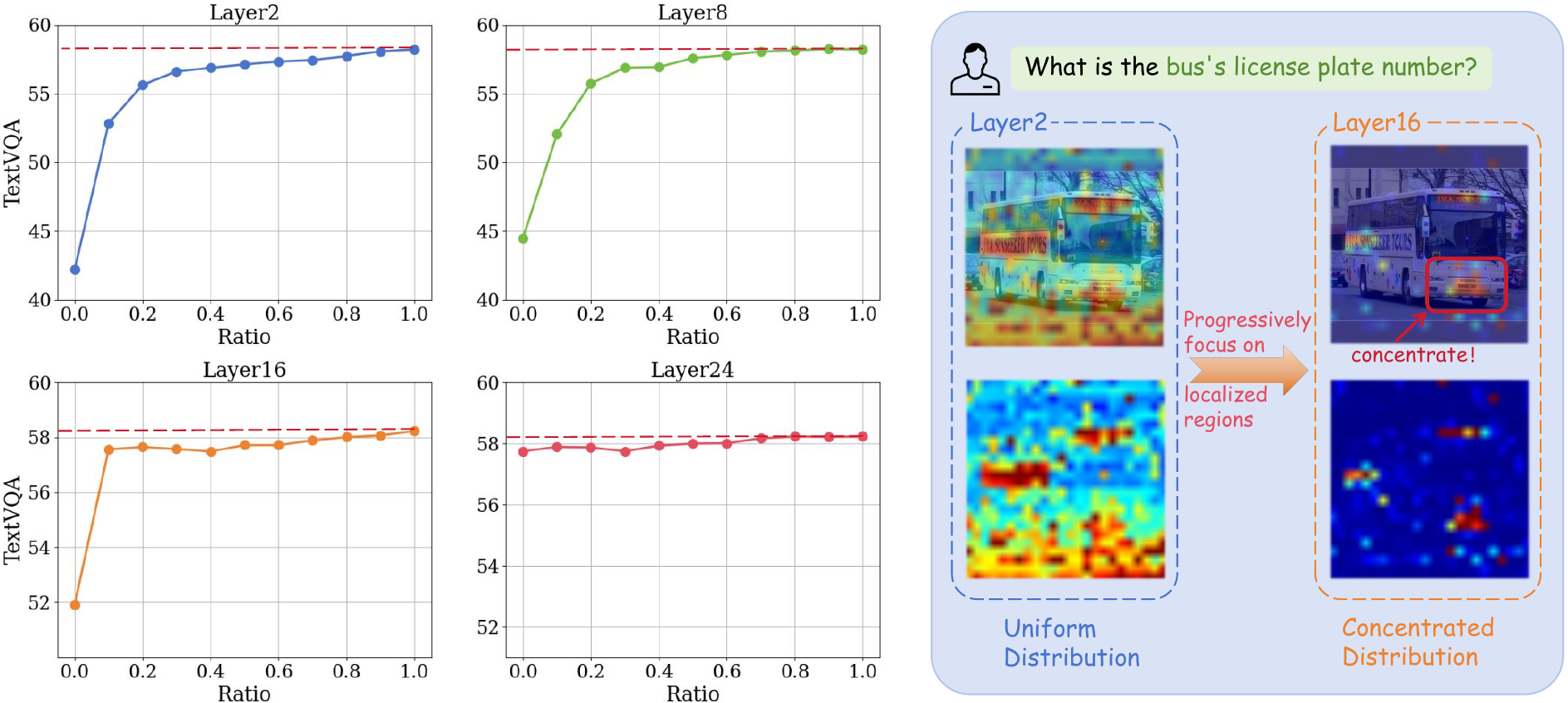}
\caption{Observatioins about visual redundancy acoross layers. Left: TextVQA performance of LLaVA-1.5 with varying ratio of retained image tokens at different layer. The preserved image tokens are those that receive the highest attention from the text tokens. Right: Visualization of attention map in shallow and deep layers. }
\label{fig:seed_textvqa_original_motivation}
\end{figure*}

\section{Introduction}
\label{sec:intro}
In recent years, Large Vision-Language Models (LVLMs) have emerged as a central focus in deep learning research \citep{liu2024visual,Dai2023InstructBLIPTG,bai2023qwen,zhang2024internlm,chen2023shikra}. Remarkable progress have been witnessed across various application domains, including image and video understanding \citep{gpt4v,gemini}. The rapid development of LVLMs is gradually paving the way for artificial intelligence to integrate into daily life \citep{Li2023BLIP2BL,Zhu2023MiniGPT4EV,Zhang2023VideoLLaMAAI,liu2024rar}.

Despite the advancements of LVLMs, a significant challenge lies the escalating computational costs. Images or videos, as continuous and information-rich signals, exhibit substantial spatial redundancy but are difficult to compress losslessly. It results in excessive vision tokens and a steep increase in training and inference costs, which becomes particularly pronounced with higher image resolutions~\citep{zhang2024internlm,wang2024qwen2,hu2024mplug} and longer videos~\cite{chen2024sharegpt4video,lin2023video,maaz2023video}. The number of vision tokens increases quadratically with the resolution or the frame numbers, driving the sequence length into the tens of thousands~\citep{li2023otterhd}. Given that the computational complexity of transformers scales with sequence length, the associated computational costs become prohibitively high~\citep{liu2024world,xu2024llava}. Consequently, there is a pressing need to reduce the redundancy and concentrate more on valuable visual information for efficient deployment. 

Previous exploration of reducing image tokens could be roughly divided into two categories: One is to compress the vision tokens before passing them into the base LLM of LVLMs~\citep{shang2024llava,arif2024hired,li2023blip,yao2024deco}.
The other is to partially drop the vision tokens at the very shallow layer of the LVLMs~\citep{chen2024image}. However, both ideas inevitably hurt the performance of LVLMs: the former suffers from the information loss introduced by their compression, and the latter drops part of the information before the LVLMs fully understand them. 

To break through these limitations, we explore the nature of LVLMs in understanding images from an intuitive question: \textit{Are all image tokens necessary for all LVLM layers?} We conduct an empirical study by removing different ratios of image tokens at different layers of the LVLM at inference time and observing the benchmark performance change. 
As shown in Figure \ref{fig:seed_textvqa_original_motivation}, the LVLMs are sensitive toward token dropping on shallow layers, regardless of the dropping ratio. However, in deeper layers, image tokens gradually become less critical to the final results. The results indicate that the LVLMs understand the image layer-by-layer and the redundancy within image tokens increases correspondingly. We further visualize the attention between the instructions and the image tokens, and observe a consistent phenomenon that in shallow layers, the LVLMs pay attention to most image tokens to understand the image globally. With the layer increasing, it tends to focus on the few tokens that are related to the instruction and the rest are unnecessary.

Based on the observation, we introduce PyramidDrop, a simple yet effective image token reduction strategy for LVLMs to accelerate both inference and training without performance loss. 
PyramidDrop divides the LVLM into several stages, dropping a portion of the image tokens at the end of each stage according to a predefined ratio. We employ a lightweight attention module to rank the image tokens and finally keep important visual concentration, which incurs negligible overhead. With this design, we retain all image tokens in the shallow layers to avoid information loss, while progressively reducing the number of tokens as the layers deepen to maximize training and inference efficiency.

Extensive experiments verify the effectiveness and efficiency of our PyramidDrop. For example, applying PyramidDrop to LLaVA-NeXT-7B~\citep{liu2024llavanext} could achieve 40\% training time reduction without sacrificing performance across 16 Vision-Language tasks. Moreover, PyramidDrop enables the LLaVA-NeXT model to be trained with doubled input resolution with only 70\% training time of the vanilla LLaVA-NeXT, and reaches a better performance on high-resolution benchmarks like DocVQA~\citep{mathew2021docvqa} and InfoVQA~\citep{mathew2022infographicvqa}. Furthermore, PyramidDrop can function as a plug-and-play strategy for inference acceleration, offering enhanced model performance and fewer FLOPs than FastV~\citep{chen2024image}.

\section{Related Work}
\paragraph{Token Reduction}
The large language model (LLM) realm has made several efforts in applying token reduction for inference acceleration and KV cache compression\citep{han2023lm}. StreamLLM\citep{xiao2023efficient} only keeps attention sinks and the most recent tokens to reduce the size of the KV cache. FastGen\citep{ge2023model} introduces an adaptive KV cache management approach that optimizes memory usage by adjusting retention strategies according to the specific properties of attention heads. Heavy-Hitter Oracle (H2O)\citep{zhang2024h2o} employs a strategy that selectively prunes key-value pairs (KVs) during generation, utilizing a scoring mechanism driven by cumulative attention to inform the removal process. ScissorHands\citep{liu2024scissorhands} concentrates on identifying and retaining important tokens that show a consistent pattern of attention weight across previous token windows during generation. These works attempt to address the redundancy of text tokens during the inference process in LLMs. As for visual tokens, existing works~\citep{liang2022patchesneedexpeditingvision,kong2022spvitenablingfastervision,cao2023pumerpruningmergingtokens,shi2024crossgetcrossguidedensembletokens,xiong2024pyraparallelyieldingreactivation} make explorations on Vision Language Models (VLMs) before the era of large vision-language models, focusing on token reduction for vision transformers (ViTs). A recent work, FastV~\citep{chen2024image}, makes an early attempt at visual token reduction in LVLMs, which drops visual tokens at the second layer of LVLMs during inference. In contrast, our work makes a more comprehensive study of the visual redundancy in LVLMs and proposes a progressive visual token reduction solution for both training and inference of LVLMs.
\paragraph{Large Vision Language Models}
Enabled by the open-sourcing of large language models like LLaMA\citep{touvron2023llama} and Vicuna\citep{chiang2023vicuna}, LVLMs\citep{chen2023pali} have advanced the ability to understand and generate diverse content by seamlessly integrating information across multiple modalities, such as text, images, and audio. Models like LLaVA\citep{liu2024visual}, InstructBLIP\citep{Dai2023InstructBLIPTG}, and MiniGPT-4\citep{zhu2023minigpt} have pushed the boundaries of this field, enabling users to interact with these intelligent systems through multimodal prompts, including images and text. Recent advances~\citep{zhang2024internlm,wang2024qwen2,hu2024mplug} have significantly increased the number of image tokens for high-resolution image understanding, resulting in substantial costs for training and inference in LVLMs. This underscores the critical importance of developing more efficient training and inference methods for LVLMs.

\section{Method}

\subsection{Study of Visual Token Redundancy in LVLMs}\label{motivation}
The fundamental design of PyramidDrop stems from an intuitive question: are all image tokens necessary for all LVLM layers? 
To explore it and reveal the nature of LVLMs, we conduct a two-variable experiment by removing different ratios of image tokens at different layers of the LVLM at inference time and observing the benchmark performance change. 

In detail, we select LLaVA-v1.5-7B~\citep{liu2024visual} as the base model, and employ a popular LVLM benchmark, TextVQA~\citep{singh2019towards}, as the evaluation data. 
TextVQA consists of a substantial number of images that contain fine-grained information like text. 
The questions in TextVQA focus on the textual elements within images, requiring LVLMs to capture the global image information while mining the great detailed visual clues. 
This characteristic increases the model's sensitivity to image token compression, enabling a more precise evaluation of redundancy.  
 
Considering LLaVA-v1.5-7B consists of 32 layers, we drop varying proportions of image tokens during inference at layer 2, 8, 16, and 24 to assess redundancy at different layers. 
The ranking of tokens is based on the attention values of text tokens towards image tokens, with the retained image tokens corresponding to those with the highest attention values. As illustrated in Figure~\ref{fig:seed_textvqa_original_motivation} (left), at layer 2, the LVLMs are sensitive toward token dropping on shallow layers, regardless of the dropping ratio. This indicates most of the image tokens in shallow layers play an important role in providing information for answering the instruction. With the layer increases, the redundancy of image tokens increases rapidly. At layer 16,  even preserving only 10\% of image tokens will not cause an obvious performance decline. Notably, at layer 24, the model performance is nearly irrelevant to the image tokens, indicating that the model has already captured the necessary image information and the image tokens are redundant for the model now.

We further validate our hypothesis with an attention map comparison between different layers. As shown in Figure~\ref{fig:seed_textvqa_original_motivation} (right), the LVLM pays attention to most of the image tokens at shallow layers and the attention to different tokens shows a uniform pattern. On the contrary, at the middle of the LVLMs, the attention shows a sparse pattern and mainly focuses on the question related image local parts.

\begin{figure*}[t]
\centering
\includegraphics[width=1.0\textwidth]{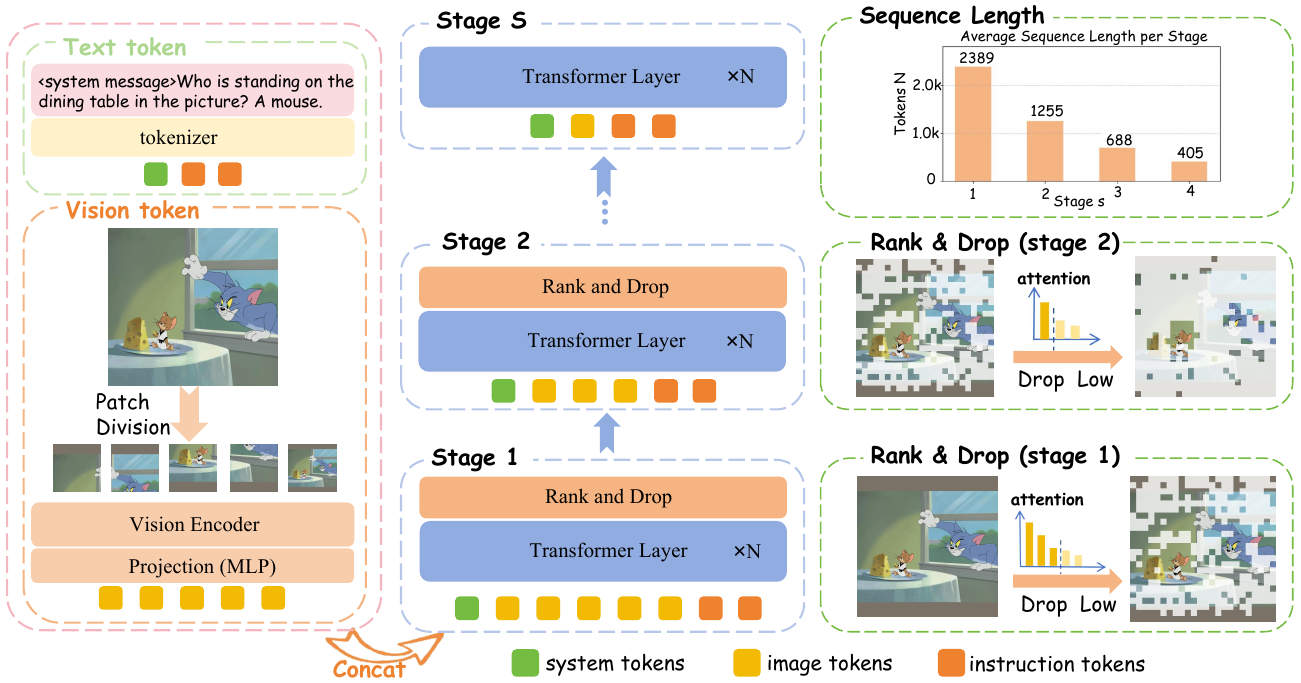}
\caption{Overview of PyramidDrop. We divide the forward pass of the LLM into multiple stages, and drop part of the image tokens at the end of each stage with a pre-defined ratio. The dropping is based on a lightweight attention calculation with a negligible time overhead, and according to this criterion, the LLM accurately selects important image tokens related to instruction. Due to the efficient redundancy reduction strategy, the average sequence length decreases rapidly.}
\label{fig:pipeline}
\end{figure*}

\subsection{PyramidDrop}
Previous research on image token compression drops image tokens before passing them to the language model or uses a fixed compression ratio across all language model layers. However, as we analyzed in Sec \ref{motivation}, redundancy is not consistent across different layers. Redundancy of image tokens is relatively minimal in the shallow layers and becomes progressively larger in deeper layers. Thus, uniformly compressing image tokens across layers may lead to the loss of valuable information in the shallow layers while retaining unnecessary redundancy in the deeper layers.

Inspired by this observation, we propose PyramidDrop, which fully leverages layer-wise redundancy to compress image tokens and finally keep important visual concentration. The pipeline of the proposed PyramidDrop is illustrated in Figure \ref{fig:pipeline}. To maximize training efficiency while preserving the essential information of the image tokens, we choose to divide the forward pass of the LLM into multiple stages. In the shallow layers, we retain a higher proportion of image tokens to preserve the entire vision information. At the end of each stage, we partially drop the image tokens, until nearly all the image tokens being eliminated in the deeper layers. This approach allows us to optimize training efficiency while maintaining critical information.

\paragraph{LVLM Pre-fill Formulation.}
We denote the vision encoder as $\mathcal{V}$, the vision-language projector as $\mathcal{P}$, the language model as $\mathcal{L}$, a pretrained LVLM as $\mathcal{M} = (\mathcal{L},\mathcal{V}, \mathcal{P} $), where $\mathcal{L} = (\mathcal{L}_0, \mathcal{F})$.  The language model consists of tokenizer $\mathcal{L}_0$ and $J$-layer transformer decoder $\mathcal{F}$. We formulate an image-text pair as $(\mathcal{V}, \mathcal{T})$, where the text is composed with an instruction and an answer $\mathcal{T}=\{T_i; T_a\}$\footnote{Here we omit the system prompt and chat format for illustrative purposes}. 
The input of the transformer $\mathcal{F}$ contains both the image tokens $v_0 = \mathcal{P}(\mathcal{V}(v))$ and the text tokens $t_0 = \mathcal{L}_0 (T)$.

During the forward pass of tokens, we can obtain the hidden states $v_j$, $t_j $ of vision tokens and text tokens in layer $j$, formally: 
\begin{equation}
    v_j, t_j = \mathcal{F}_j(v_{j-1}, t_{j-1}) 
\label{eq:eq1}
\end{equation} 

\paragraph{Progressive Visual Redundancy Reduction.}
We partition the language into $\mathcal{S}=\{s_n\}_{n=0}^S$ stages, and remove the image tokens $v$ with a pre-defined ratio $\lambda$ at the end of each stage. Formally, with the image tokens $v_{s_n}$ as the input of stage $s_n$, we remove $\lceil (1-\lambda) \cdot  |v_{s_n}| \rceil$ tokens from the $v_{s_n}$ and treat the rest image tokens as the next stage input $v_{s_{n+1}}$.

Following our observation in Sec 3.1, the attention value between image and text tokens could reflect the image token importance properly, so we based on it to realize the drop operation. With the concern of calculation efficiency and training-inference consistency, we calculate the attention between all the image tokens and the last token of the instruction (we denote it as $t_j^I$, the last-instruction token in the following).  

Formally, we denote the last layer of stage $s_n$ as $F_j$, we obtain key states of the image tokens as $k_j^{v}$ and the query state of last instruction token $q_j^{t_{I}}$ with the following operation:
\begin{equation}
    k_j^{v} = \mathcal{K}_j(v_j), \quad q_j^{t_{I}} = \mathcal{Q}_j(t_j^I).
\label{eq:eq1}
\end{equation}
where $\mathcal{Q}_j$, $\mathcal{K}_j$ are the query matrix and the key matrix reused from the self-attention block of $F_j$.

We calculate the similarity with $q_j^{t_{I}} \times (k_j^{v})^T$ and drop part of the image tokens based on the drop ratio $\lambda$. The image token number decreases exponentially stage by stage, and close to zero in the deeper layers. We denote the image token number of $v_0$ as $V = |v_0|$, and the image token number at each stage $V_s$ could be calculated as:

\begin{equation}
    V_s =  V_0 \cdot \lambda^{s-1}, \quad s = 1,2,\dots,S
\end{equation}


\setlength{\tabcolsep}{5pt}
\begin{table*}[!t]
\renewcommand{\arraystretch}{1}
\small
\centering
\vspace{1pt}
\begin{tabular}{@{\extracolsep{\fill}}c|cc|cccccccc@{}} 
 \toprule[1.2pt]
  Model & \makecell{Inference\\Strategy} & TFLOPS  & MME   & SQA$^I$    & MMB$^{CN}$ & GQA & POPE & \makecell{TextVQA} &  \makecell{SEED$^I$}  & Avg\\ \midrule
      
\multirow{3}{*}{\makecell{LLaVA-NeXT-7B}}   
 ~& vanilla & 20.8 & 1534.1 & 70.4 & 60.5 & 64.2 & 86.1 & 67.2 & 71.1 & 70.9  \\ 
 ~& FastV & 10.6 & 1504.0 & 69.3 & \textbf{60.0} & 63.5 & 86.3 & 66.5 & 69.3 & 70.1  \\
 ~& \cellcolor[HTML]{E2E2E2}PDrop & \cellcolor[HTML]{E2E2E2}9.5 & \cellcolor[HTML]{E2E2E2}\textbf{1533.0} & \cellcolor[HTML]{E2E2E2}\textbf{69.4} & \cellcolor[HTML]{E2E2E2}59.9 & \cellcolor[HTML]{E2E2E2}\textbf{63.9} & \cellcolor[HTML]{E2E2E2}\textbf{86.4} & \cellcolor[HTML]{E2E2E2}\textbf{67.0} & \cellcolor[HTML]{E2E2E2}\textbf{70.0}  & \cellcolor[HTML]{E2E2E2}\textbf{70.5} \\
 \midrule

\multirow{4}{*}{\makecell{LLaVA-1.5-7B}}   
 ~& vanilla & 3.82 & 1510.7 & 66.8 & 58.3 & 62 & 85.9 & 58.2 & 66.1 & 67.5 \\ 
 ~& FastV & 2.01 & 1473.7 & 68.5 & 57.3 & 59.4 & 84.0 & 57.2 & 64.0 & 66.4  \\
 ~& \cellcolor[HTML]{E2E2E2}PDrop & \cellcolor[HTML]{E2E2E2}1.78 & \cellcolor[HTML]{E2E2E2}\textbf{1500.8} & \cellcolor[HTML]{E2E2E2}\textbf{69.2} & \cellcolor[HTML]{E2E2E2}\textbf{58.5} & \cellcolor[HTML]{E2E2E2}\textbf{60.1} & \cellcolor[HTML]{E2E2E2}\textbf{84.8} & \cellcolor[HTML]{E2E2E2}\textbf{57.6} & \cellcolor[HTML]{E2E2E2}\textbf{64.3} & \cellcolor[HTML]{E2E2E2}\textbf{67.1} \\ 

\bottomrule[1.2pt]
\end{tabular}
\caption{\textbf{Inference acceleration performance}. We compare PyramidDrop, FastV and vanilla model, and find PyramidDrop outperforms FastV on almost all benchmarks. PyramidDrop here is as an inference-only strategy for LVLMs. The highest score is denoted in \textbf{bold}.}
\vspace{-5pt}
\label{table:inference compare pdrop}
\end{table*}

\begin{table*}[!t]
    \small
    \centering
    \setlength{\tabcolsep}{3mm}{
    \begin{tabular}{lc|ccccccc}
        \toprule
        Method & \makecell{Average\\tokens} & MME & MMB & SQA & GQA &TextVQA & Average & Ratio
        \\
        \midrule
        LLaVA-1.5-7B & 576 & 1862 & 64.7 & 69.5 & 61.9 & 58.2 & 69.4 & 100\%
        \\
        \midrule
        ToMe & 192 & 1563 & 60.5 & 65.2 & 54.3 & 52.1 & 62.0 & 89.9\%
        \\
        FastV & 192 & 1612 & 61.2 & 67.3 & 52.7 & 52.5 & 62.9 & 90.6\%
        \\
        SparseVLM & 192 & 1721 & 62.5 & 69.1 & \textbf{57.6} & 56.1 & 66.3 & 95.5\% \\
        \rowcolor{gray!20}
        PDrop & 192 & \textbf{1797} & \textbf{63.3} & \textbf{69.2} & 57.3 & \textbf{56.5} & \textbf{67.2} & \textbf{96.8\%}
        \\
        \midrule
        ToMe & 128 & 1343 & 53.3 & 59.6 & 52.4 & 49.1 & 56.3 & 81.1\%
        \\
        FastV & 128 & 1490 & 56.1 & 60.2 & 49.6 & 50.6 & 58.2 & 83.9\%
        \\
        SparseVLM & 128 & 1696 & 60.0 & 67.1 & 56.0 & 54.9 & 64.6 & 93.0\% \\
        \rowcolor{gray!20}
        PDrop & 128 & \textbf{1761} & \textbf{61.6} & \textbf{68.4} & \textbf{57.1} & \textbf{56.6} & \textbf{66.4} & \textbf{95.6\%}
        \\
        \midrule
        ToMe & 64 & 1138 & 43.7 & 50.0 & 48.6 & 45.3 & 48.9 & 70.5\%
        \\
        FastV & 64 & 1256 & 48.0 & 51.1 & 46.1 & 47.8 & 51.2 & 73.7\%
        \\
        SparseVLM & 64 & 1505 & 56.2 & 62.2 & 52.7 & \textbf{51.8} & 59.6 & 85.9\% \\
        \rowcolor{gray!20}
        PDrop & 64 & \textbf{1561} & \textbf{58.8} & \textbf{69.0} & 47.5 & 50.6 & \textbf{60.8} & \textbf{87.6\%}
        \\

        \bottomrule
    \end{tabular}
    }
    \caption{\textbf{Compare PyramidDrop with other efficient inference strategies with different image tokens.} By retaining an average of 192, 128, and 64 image tokens, PyramidDrop achieves sota results, demonstrating its ability to deliver optimal performance at lower compression ratios. Furthermore, even as the compression ratio increases, PyramidDrop maintains robust performance, highlighting its strong resilience. The design of Conical Visual Concentration maximizes efficiency without compromising performance. \textbf{PyramidDrop is an inference-only method here.}}
    \label{tab:infer comparison}
\end{table*}

\paragraph{Efficiency Analysis of PyramidDrop}
Here we analyze the efficiency from two parts: the computation overhead introduced by PyramidDrop, and the input sequence computation cost economized by PyramidDrop.

The extra computation cost introduced by PyramidDrop mainly lay in the similarity computing for image token ranking. Benefiting from our design, the calculation is only between a query toke and $V_s$ image tokens, so its computation complexity is $O(n)$ and only $S-1$ times in the forward process. 
Further, we notice the importance of FalshAttention in practice, so we keep using it during training and extract the query and key token from the original forward to calculate our lightweight similarity matrix. 
 
When it comes to the computation cost economized by PyramidDrop. With the consideration of FlashAttn~\citep{dao2022flashattentionfastmemoryefficientexact}, we roughly define the forward inference cost of a layer with $N$ image tokens as a linear function with a constant factor $c$ that $ c\cdot L$, so the overall computation cost of an LVLM with $L$ layers is $c\cdot N\cdot L$. When using PyramidDrop with S stages and the ratio $\lambda$, the overall computation cost is:
\begin{equation}
    \frac{1-\lambda^S}{S\cdot(1-\lambda)} \cdot c \cdot N  \cdot L
\end{equation}
For example, if $\lambda=0.5$ and we reduce the redundancy with 4 stages, it could save nearly $53.2\%$ computation cost theoretically, and we find this setting has a neglectable performance influence for models in practice.

\section{Experiment}
\subsection{Setup}

\setlength{\tabcolsep}{1.5pt}
\begin{table*}[!t]
\renewcommand{\arraystretch}{1}
\footnotesize
\centering
\resizebox{1.0\textwidth}{!}{
\begin{tabular*}{\textwidth}{@{\extracolsep{\fill}}c|ccccc|ccccccccc@{}}
 \toprule[1.2pt]
  Model & Train \& Infer & \makecell{\#Patch} & \makecell{GPU\\hours}  &\makecell{Reduced\\Training Time}& \makecell{Infer\\Flops(T)}  & MME  & MMB & MMB$^{CN}$  & SEED$^I$ & \makecell{MM\\Star}  & POPE & SQA$^I$ & AI2D& Avg\\ \midrule
      
\multirow{4}{*}{\makecell{LLaVA\\-NeXT-7B}} ~& vanilla & 5  & 366 & 0\%& 20.8 & 1534.1 & 68.7 & 60.5  & 71.1 & 41.1   & 86.1 & 70.4& 66.1 & 67.6 \\
 ~& PDrop & 5 & 218 &40.4\% & 9.46 & 1540.8 & 67.8 & 60.6  & 69.9 & 41.7   & 86.5 &70.1 &66.7 &67.5 \\
 \cmidrule{2-15}
 ~& vanilla & 9 & 483  & 0\% & 40.6 & 1544.7 & 67.4 & 60.0  & 69.5 & 40.0   & 86.3 &68.8 &65.0 & 66.8 \\
 ~& PDrop & 9 & 269 & 44.3\% & 18.1 & 1542.0 & 68.1 & 61.0  & 70.3 & 40.9  & 86.6 & 69.4&66.1 &67.4  \\
 \midrule
 
\multirow{2}{*}{\makecell{LLaVA\\-1.5-7B}} ~& vanilla & 1 & 104 & 0\% & 3.82 & 1510.7 & 64.3 & 58.3 & 66.1 & 33.2  & 85.9 & 66.8& 55.6& 63.2\\
 ~& PDrop & 1 & 79 & 24.0\%  & 1.78 & 1467.3 & 66.1 & 58.5 & 65.5   & 34.0 & 86.0 & 71.0&56.5 & 63.9\\ 
\bottomrule[1.2pt]
\end{tabular*}
}
\caption{\textbf{PyramidDrop greatly accelerate LVLM training while keeping the general multimodal abilities on 8 popular LVLM benchmarks.} ``Infer Flops'' means using PyramidDrop for the inference of PyramidDrop-trained models. ``\#Patch'' means the total number of local patches and global patch after processing a single image. Benchmark
names are also abbreviated as following. MMB: MMBenchmark \citep{liu2023mmbench}; MMB$^{CN}$: MMBench-Chinese \citep{liu2023mmbench}; SEED$^{I}$: SEED-Bench (Image) \citep{li2023seed}; SQA$^I$:ScienceQA-IMG\citep{lu2022learn}.}
\label{table:main table}
\end{table*}

\setlength{\tabcolsep}{1.5pt}
\begin{table*}[!t]
\renewcommand{\arraystretch}{1}
\footnotesize
\centering
\begin{tabular*}{\textwidth}{@{\extracolsep{\fill}}c|ccccc|ccccccccc@{}}
 \toprule[1.2pt]
  Model & Train \& Infer  & \makecell{\#Patch} & \makecell{GPU\\hours} &\makecell{Reduced\\Training Time} &\makecell{Infer\\Flops(T)} & \makecell{DocVQA}  & \makecell{InfoVQA} & \makecell{TextVQA} & \makecell{ChartQA} & \makecell{OCRVQA} & \makecell{VQAV2} & \makecell{VizWiz} & GQA & Avg\\ \midrule
      
\multirow{4}{*}{\makecell{LLaVA\\-NeXT-7B}}  ~&  vanilla & 5 & 366 &0\% &20.8 & 70.0 & 33.3 & 67.2 & 64.0 & 63.7 & 81.7 & 59.6 & 64.2 & 63.0   \\
  ~& PDrop & 5 & 218 &40.4\% &9.46 & 69.0 & 31.7 & 67.7 & 63.1 & 63.1 & 81.5 & 61.0 & 63.9 & 62.6   \\
  \cmidrule{2-15}
  ~&  vanilla& 9  & 483 &0\%  &40.6& 74.3 & 36.2 & 67.6 & 63.0 & 63.8 & 81.6 & 58.0 & 63.5 &  63.5  \\
  ~& PDrop &9 &269  &44.3\% & 18.1&75.0 &37.4 &68.4 &64.3 &63.5 &81.7 &60.6 &64.1 &64.4   \\ 

\bottomrule[1.2pt]
\end{tabular*}
\caption{\textbf{PyramidDrop greatly accelerate LVLM training while keeping abilities on other 8 high-resolution benchmarks}. ``Infer Flops'' means using PyramidDrop for the inference of PyramidDrop-trained models. We report more benchmarks which contain lots of fine-grained visual information. }
\vspace{-8pt}
\label{table:table:main table, higher resolution}
\end{table*}

\textbf{Models} We verify the effectiveness and generalization of the proposed PyramidDrop by experiment on LVLMs with different architectures and input resolution. In detail, we study LLaVA-1.5-Vicuna-7B \citep{liu2024visual}, LLaVA-NeXT-Vicuna-7B \citep{liu2024llavanext}.  LLaVA-1.5 is the most widely used open-source LVLM backbone for research, which is designed with a simple yet effective architecture that maps the 576 image features from the CLIP encoder as the LLM input with a projector. LLaVA-NeXT is the high-resolution extension of LLaVA-1.5, which supports at most 2880 image tokens and has better high-resolution capability. 

\noindent\textbf{Benchmarks}
To thoroughly evaluate our image token compression strategy, we conduct experiments across 16 benchmarks. 
The MME Benchmark \citep{fu2023mme} assesses the perception and cognitive abilities of LMMs. MMBench and MMBench-CN \citep{liu2023mmbench} are benchmarks that manually craft questions to evaluate vision-related reasoning and perception in both English and Chinese, respectively. SEED \citep{li2023seed}, generated with the aid of GPT-4, comprises a dataset of approximately 19,000 questions pertaining to images and videos. MM-Vet \citep{yu2023mm} leverages GPT-4 for a six-dimensional evaluation of LMM capabilities. In the realm of traditional VQA benchmarks, such as VQA-v2 \citep{goyal2017making} and VizWiz \citep{gurari2018vizwiz}, are also utilized. Additionally, several benchmarks featuring higher-resolution visual content, including DocVQA \citep{mathew2021docvqa}, ChartQA \citep{masry2022chartqa}, InfographicVQA \citep{mathew2022infographicvqa}, and TextVQA \citep{singh2019towards}. Finally, MMStar \citep{chen2024we} presents tasks with strong visual dependency, minimal data leakage, and requires sophisticated multimodal capabilities.

\noindent\textbf{Efficientness Evaluation}  
We consider both the training time efficiency evaluation and inference time throughout. For training efficiency, we report the real training GPU hours with the same devices. For inference throughout, we follow the FastV\citep{chen2024image} and report the FLOPs of the image token part. In detail, we consider the FLOPs of the multi-head attention and the feed-forward network modules as \(4nd^2 + 2n^2d + 2ndm\), where \(n\) is the number of tokens, \(d\) is the hidden state size, and \(m\) is the intermediate size of the FFN. Considering there are three linear layers in FFN of LLaMA, the FLOPs is modified as \(4nd^2 + 2n^2d + 3ndm\). Our PyramidDrop has different image token numbers at different stages and the FLOPS could be calculated by:

\begin{table*}[!t]
    \small
    \centering
    \begin{tabular}{l|ccc|ccccccccc}
        \toprule
        Method & \makecell{Average\\tokens} & \makecell{GPU\\hours} & \makecell{Infer\\Flops(T)} & POPE & SQA & MMB & GQA & \makecell{OCR\\VQA} & SEED$^I$ & MMStar & AI2D & \makecell{Text\\VQA} 
        \\
        \midrule
        LLaVA-1.5-7B & 576 & 104 (100\%) & 3.82 & 85.9 & 66.8 & 64.3 & 62.0 & 59.8 & 66.1 & 33.2 & 55.6 & 58.2 
        \\
        \midrule
        Q-former & 288 & 88 (84.6\%) & 1.89 & 67.2 & 66.9 & 53.8 & 41.3 & 19.0 & 49.2 & 28.6 & 51.8 & 44.4  
        \\
        FastV & 306 & 81 (78.0\%) & 2.01 & 85.2 & 69.5 & 65.6 & 61.0 & 60.7 & 65.3 & 33.4 & 55.3 & 58.4 
        \\
        LLaVolta & 339 & 93 (89.4\%) & 3.82 & 85.6 & 69.6 & 63.6 & \textbf{62.2} & 60.0 & \textbf{66.3} & 33.2 & 55.7 & 58.3  \\
        \rowcolor{gray!20}
        PDrop & \textbf{270} & \textbf{79 (76.0\%)} & \textbf{1.78} & \textbf{86.0} & \textbf{71.0} & \textbf{66.1} & 61.9 & \textbf{61.0} & 65.5 & \textbf{34.0} & \textbf{56.5} & \textbf{58.5} 
        \\

        \bottomrule
    \end{tabular}
    \caption{\textbf{Compare PyramidDrop with other efficient training strategies.} Average tokens here refer to the average image tokens across all LLM layers, while GPU hours represent the time required for model training. As shown in the table, our method achieves the best performance on nearly all benchmarks while also being the most cost-effective strategy in terms of both training and inference.}
    \label{tab:training comparison}
\end{table*}

\begin{equation}
\begin{aligned}
\sum_{s=0}^{S-1} K_{s} \left( 4n_{s}d^2 + 2n_{s}^2d + 3n_{s}dm \right) \\
\text{s.t.} \quad n_s = \lambda^{s} n, \quad s = 0,1,2,\dots,S-1
\end{aligned}
\end{equation}

\noindent\textbf{Implementation details}
Given that the LLM within the LVLM used in our experiments consists of 32 layers, we employ a straightforward approach by fixing \( S \) to 4, effectively dividing the LLM into four equal parts. This segmentation allows the forward pass to be divided into four stages, with the number of image tokens decreasing exponentially at each stage. During accelerated training, we can adjust the value of \( \lambda \) to control the proportion of image tokens that are pruned, and by default, \( \lambda = 0.5 \). We conduct all the experiments on 8 NVIDIA A100 80GB GPUs.

It is important to note that, we apply FlashAttn~\citep{dao2022flashattentionfastmemoryefficientexact} during both training and inference as we don't need to output full attention map. And since the LLaVA-NeXT model's data and training code are not open-source, we conduct training based on the open-source project Open-LLaVA-NeXT \citep{chen2024open}. Due to differences in a portion of the training data, the benchmark performance may vary compared to that of LLaVA-NeXT~\citep{liu2024llavanext} blog. 

\subsection{Efficiency of PyramidDrop in Inference}
\paragraph{PyramidDrop outperforms SOTA methods as a inference-only strategy.}
As illustrated in Table~\ref{table:inference compare pdrop}, we directly apply the multi-stage compression strategy during the inference phase of the vanilla model, comparing it with the inference acceleration approach, FastV. The results on LLaVA-Next demonstrate that our method outperforms FastV across various critical benchmarks. Specifically, we achieve an impressive score of 1533.0 on MME, surpassing Fastv by 1.5\%, while also exceeding it by 0.4\% on GQA. Notably, the advantages of our method is also pronounced in high-resolution benchmarks. For instance, on the relatively challenging TextVQA, our approach outperforms FastV by 0.5\%, and on SEED-Bench (Image), we achieve improvements of 0.7\%.

Results from LLaVA-1.5 reveal similar trends across multiple benchmarks, including MME, ScienceQA, and MMBenchCN, where our method not only demonstrates superior performance but also achieves a greater reduction in FLOPs. When compared to the baseline, our approach consistently reaches comparable performance levels across most benchmarks, while effectively mitigating information loss in high-resolution benchmarks. 
These findings indicate that FastV’s premature compression of image tokens leads to inevitably image information loss and significant performance declines in many benchmarks, whereas our multi-stage compression strategy preserves critical information from image tokens while maximizing the elimination of redundancy.
The observation is also consistent with our finding in Sec~\ref{motivation} that in shallow layers, most image tokens are critical for LVLMs to understand the image properly, while in deep layers, most of them are redundant for LVLMs.
We also compare PyramidDrop with three baseline methods: ToMe\citep{bolya2022token}, FastV, and SparseVLM~\citep{zhang2024sparsevlm} in Table~\ref{tab:infer comparison} with different image tokens.

\begin{figure*}[t]
\centering
\includegraphics[width=1.0\textwidth]{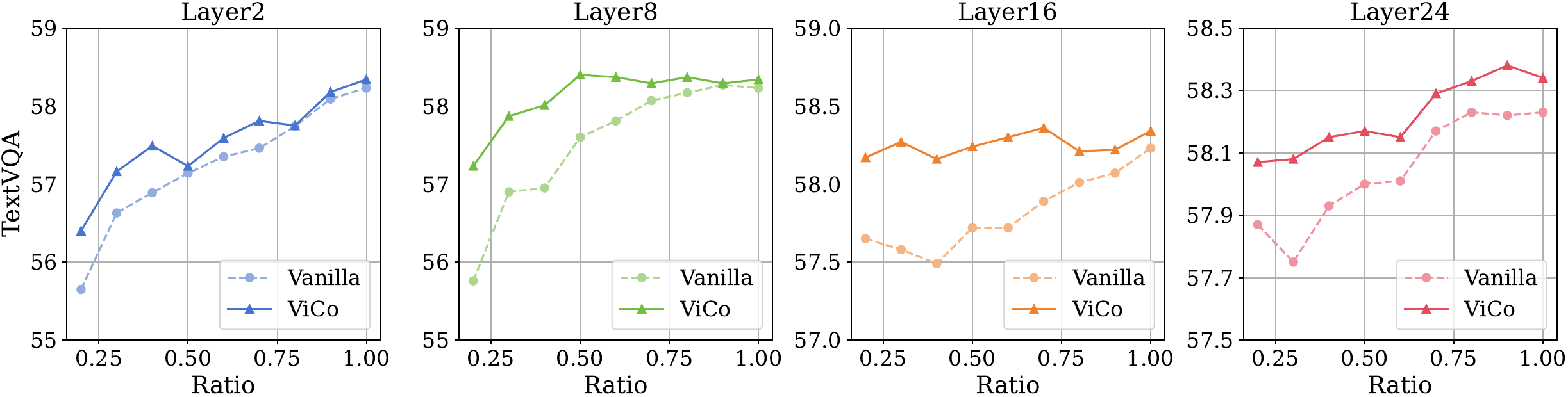}
\caption{LVLMs trained by PyramidDrop can condense key visual information into fewer vision tokens. We compare the performance of the vanilla and PyramidDrop-trained LLaVA-1.5 models, where we preserve different ratios of image tokens at layer 2, 8, 16, and 24, respectively. The horizontal axis represents the proportion of retained image tokens according to attention score. }
\vspace{-5pt}
\label{fig:comapare pdrop and original}
\end{figure*}
\begin{table}[t]
    \centering
    \resizebox{1.0\columnwidth}{!}{
    \begin{tabular}{cc|cc|cc|cc|cc}
    \toprule
        \multirow{2}{*}{Model} &
        \multirow{2}{*}{\makecell{TFLOPS}} &
        \multicolumn{2}{c|}{TGIF} & \multicolumn{2}{c|}{MSVD} & \multicolumn{2}{c|}{MSRVTT} & \multicolumn{2}{c}{Avg} \\ 
        ~&&Acc & Score & Acc &  Score & Acc & Score & Acc & Score\\
        \midrule
        Video-LLaVA &14.4 & 47.0 & 3.34 & 69.7 &3.90 & 57.8 & 3.48 &58.1 & 3.57   \\ 
        \quad w/ FastV &7.4 & 47.6 & 3.35 & 70.3 & 3.92 & 57.4 & 3.47 &58.4 & 3.58  \\
        \quad w/ PDrop &6.6 & 46.9 & 3.35 & 70.0 & 3.92 & 58.0 & 3.50 &57.9 & 3.56  \\

        \bottomrule
    \end{tabular}}
    \caption{Inference acceleration on video-LLMs. GPT-Evaluation Results on Video Question Answering Tasks are reported. We apply PyramidDrop as an inference-only strategy to vanilla Video-LLaVA.}
    \label{tab:video_result_infer}
    \vspace{-10pt}
\end{table}

\begin{figure}[t]
\centering
\includegraphics[width=1\linewidth]{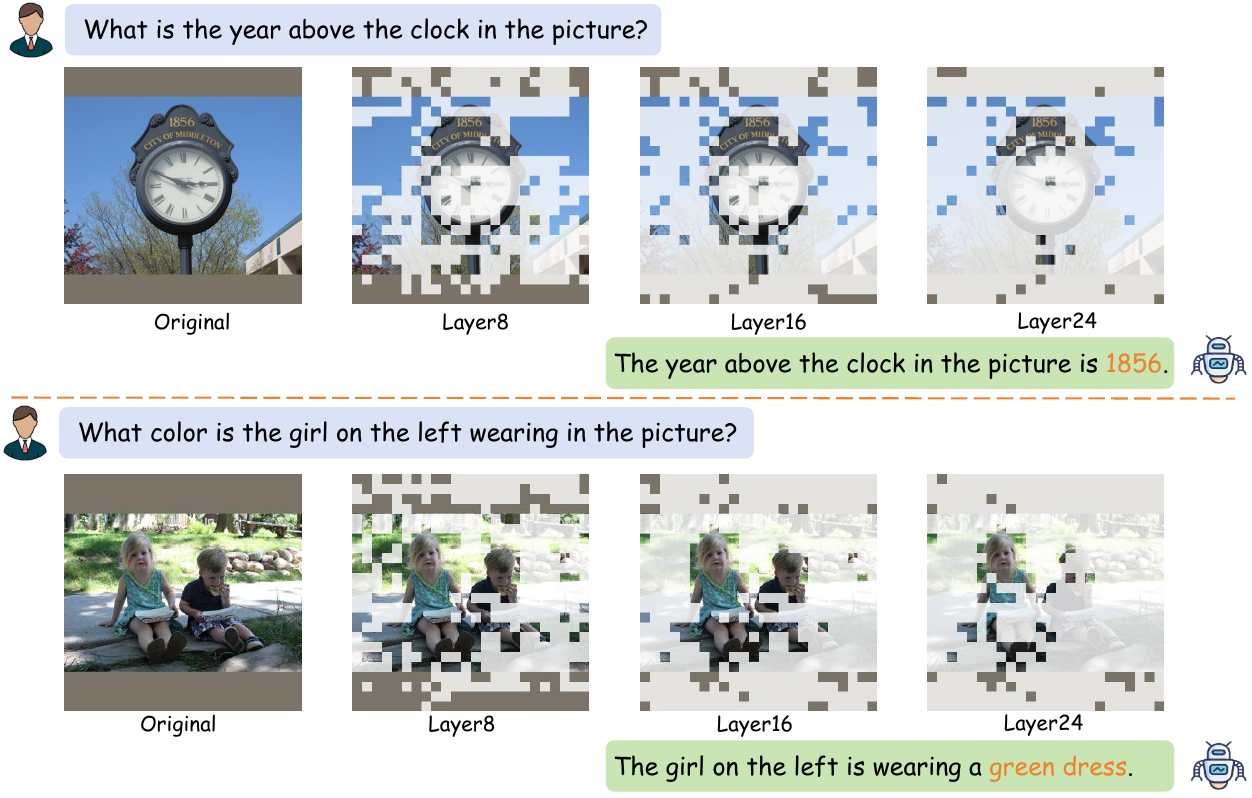}
\caption{Visualization of token dropping in LLM of LLaVA
-1.5 with PyramidDrop. PyramidDrop helps to We find LLM accurately retain image tokens according to instruction and gradually concentrate on important image patches without information loss.
}
\vspace{-10pt}

\label{fig:chosen patch}
\end{figure}

\textbf{Efficient inference on Video LLMs.}
Table~\ref{tab:video_result_infer} shows the results of using PyramidDrop as an inference-only strategy to accelerate LVLM inference. We perform zero-shot question answering on TGIF, MSVD, and MSRVTT, and the results indicate that both accuracy and score are comparable to those of the vanilla Video-LLaVA model. This demonstrates that our strategy, along with FastV, can achieve performance on par with the vanilla model. Notably, PyramidDrop achieves lower inference FLOPs by progressively eliminating redundant elements, which contributes to its efficiency. This result also suggests that the video understanding task is relatively simple, with substantial redundancy between frames. Thus, even an aggressive token-pruning strategy does not significantly impact performance, and final accuracy remains largely unaffected. In the future, further exploration is needed to improve the efficiency of video models in handling more complex visual question-answering tasks. The redundancy between frames differs significantly from that between individual images, necessitating specialized designs to effectively compress this redundancy.

\textbf{LVLM with PyramidDrop effectively preserves image tokens related to instruction.} As shown in Figure~\ref{fig:chosen patch}, we visualize the image tokens retained by LLaVA-1.5 with PyramidDrop in different stages. It is evident that when the user asks about a small object in the image, the LLM accurately identifies the region containing the relevant information based on the instructions and provides the correct answer. This demonstrates that PyramidDrop effectively leverages the LLM’s nature to understand images. The token dropping applied during inference in PyramidDrop does not lead to a loss of valuable information; on the contrary, PyramidDrop gradually selects the core patches in the image, concentrating on the most important regions. As presented in the picture, PyramidDrop helps to accurately locate big or little objects in image.

\setlength{\tabcolsep}{1.5pt}
\begin{table*}[!t]
\renewcommand{\arraystretch}{1}
\small
\centering
\vspace{1pt}
\begin{tabular*}{1.0\textwidth}
{@{\extracolsep{\fill}}c|ccccc|ccccccccc@{}}

 \toprule[1.2pt]
  Model & $\lambda$ & \makecell{GPU\\hours} &\makecell{Reduced\\Training Time} & \makecell{\#Patch} & \makecell{Infer\\Flops(T)} & MME  & MMB & GQA & MMB$^{CN}$ & SEED$^I$ & \makecell{DocVQA}& \makecell{InfoVQA} & Avg\\ \midrule
      
\multirow{4}{*}{\makecell{LLaVA-NeXT-7B}}   
 ~& vanilla & 366 & 0\% &5  & 20.8 & 1534.1 & 68.7 & 64.2 & 60.5  & 71.1  & 70.0 & 33.3 & 63.5  \\ 
 ~& 0.4 & 204 & 44.3\%& 5 & 8.22 & 1558.4 & 68.1 & 63.7 & 60.5  & 69.5  & 66.6 & 31.8 & 62.6 \\
 ~& 0.5 & 218 & 40.4\% & 5 & 9.46 & 1540.8 & 67.8 & 63.9 &60.6  & 69.9  & 69.0 & 31.7 & 62.8\\
 ~& 0.6 & 240 & 34.4\% &5  & 11.0 & 1511.4 & 68.1 & 64.1 &60.5  & 70.4  & 69.8 & 33.0 & 63.1 \\
 \midrule
 
\multirow{4}{*}{\makecell{LLaVA-1.5-7B}}      
 ~& vanilla & 104 & 0\% &1 & 3.82 & 1510.7 & 64.3 & 62.0 & 58.3 & 66.1 & 21.4 & 20.4 & 52.6 \\ 
 ~& 0.4 & 75  &27.8\% & 1 & 1.54 & 1478.8 & 66.2 & 61.7 & 58.0 & 64.5 & 21.1 &19.9 & 52.2 \\
 ~& 0.5 & 79  &24.0\% &1 & 1.78 & 1467.3 & 66.1 & 61.9 & 58.5 & 65.5 & 21.5 & 20.2 & 52.4 \\ 
 ~& 0.6 & 82  &21.1\% & 1 & 2.06 & 1471.8 & 65.9 & 62.0 & 58.9 & 65.1 & 22.5 & 21.0 & 52.7\\ 

\bottomrule[1.2pt]
\end{tabular*}
\caption{Ablation study results about $\lambda$. $\lambda$ balances the performance and efficiency of PyramidDrop, a larger $\lambda$ preserves more image information but slows down the training, and a smaller $\lambda$ has higher speedup while may influence the model performance. We adjust $\lambda$ form 0.4 to 0.6 for investigating the influence on performance and training time.}
\vspace{-5pt}
\label{table:ablation}
\end{table*}

\subsection{Efficiency of PyramidDrop in Training}

\textbf{Effective for diverse settings.}
We first study the PyramidDrop on both LLaVA-1.5 and LLaVA-Next. To further validate the effectiveness of our method, we conduct comparisons using the identical training recipe as LLaVA-1.5-7B~\citep{liu2024improved} with three other baselines: Q-Former~\citep{li2023blip}, FastV~\citep{chen2024image}, and LLaVolta~\citep{chen2024llavolta}. As shown in Table~\ref{table:main table}, PyramidDrop reduces the training time  (including both pretraining and fine-tuning stages) of the LLaVA-Next from 366 to 218 GPU hours, resulting in an impressive 40\% reduction in overall time. Besides the promising efficiency improvement, the model's performance remains comparable to the original on 16 different benchmarks. Notably, for fine-grained benchmarks like TextVQA, DocVQA, and OCRVQA, images contain a large amount of text and even documents, which request a dense and fine-grained understanding of the image. Even in this case, our approach still maintain performance at the original level. This indicates that our method successfully compresses redundant information while preserving the most critical image content.

In the case of LLaVA-1.5, which processes fewer image tokens per sample, the acceleration is not as pronounced as with LLaVA-NeXT. However, it still offers a nearly 20\% improvement in speed with comparable performance. This underscores the potential of our method to enhance training efficiency across different model configurations.

\noindent\textbf{Higher resolution at a lower cost.}
The PyramidDrop is proposed to reduce the redundancy within image tokens, and as we observed above, it enjoys higher speedup with the increase of the image/text token ratio. In this part, we explore its performance with higher image/text token ratio. 
In detail, LLaVA-NeXT is designed with a flexible image processing strategy in which an image is divided into a maximum of four local patches and a global patch, leading to at most 2880 image tokens. We denote it as LLaVA-NeXT-p5 and experiment on the LLaVA-NeXT-p9 by increasing the maximum local patches into 8 patches.  

As shown in Table~\ref{table:table:main table, higher resolution}, with the increased image/text ratio, PyramidDrop reaches a higher speedup that only 269 GPU hours is used for training, which is only 55\% of the vanilla LLaVA-Next-p9. 
Besides the superb speedup, the model trained with PyramidDrop achieves a slightly higher average performance across the 16 benchmarks. We argue too many image tokens with redundant information may confuse the LVLMs and hinder their performance, while our PyramidDrop efficiently reduce the image tokens number and helps the LVLM to focus on the critical information.
Furthermore, it is worth noting that the training time is even 70\% of the original LLaVA-Next-p5 but achieves better performance on diverse tasks, showcasing the superb efficiency and effectiveness of PyramidDrop.

\textbf{PyramidDrop training encourages compact image understanding.} 
Then we dive into the properties of the model trained with PyramidDrop and conduct experiments to investigate the changes in image token redundancy. Two models are employed for this exploration: the vanilla LLaVA-1.5 and the LLaVA-1.5 trained with our approach. As illustrated in Figure~\ref{fig:comapare pdrop and original}, we plot the TextVQA scores against the retained image tokens at layers 2, 8, 16, and 24, maintaining the same experimental settings as Sec~\ref{motivation}. We find that the curve of models trained with PyramidDrop keeps higher than the vanilla one. 
The phenomenon suggests that, for a given proportion of retained image tokens, model trained with PyramidDrop preserves more image information and achieves better performance. Alternatively, at equivalent performance levels, our method allows for a higher ratio of image tokens to compress. This improvement can primarily be attributed to the multi-stage training strategy, which progressively prunes image tokens, encouraging the model to consolidate essential information into a smaller set of tokens, resulting in more densely informative representations.

\textbf{Efficient training on Video LLMs.} Despite its success in image understanding tasks, we further investigate the efficiency of PyramidDrop in video understanding tasks. As shown in Table~\ref{tab:video_result}, applying our acceleration method on Video-LLaVA reduces the training time from 183 GPU hours to 132 GPU hours, achieving a 27.8\% reduction in training time while obtaining comparable results on the video benchmark. We perform zero-shot question answering on TGIF, MSVD, and MSRVTT, yielding relatively similar results. This outcome further underscores that our method is not only suitable for high-resolution models but also applicable to video-based vision-language models, demonstrating the broad applicability of our acceleration approach.

\begin{table}[t]

    \centering
    \resizebox{1.0\columnwidth}{!}{
    \begin{tabular}{cc|cc|cc|cc|cc}
    \toprule
        \multirow{2}{*}{Model} &
        \multirow{2}{*}{\makecell{Training GPU\\hours}} &
        \multicolumn{2}{c|}{TGIF} & \multicolumn{2}{c|}{MSVD} & \multicolumn{2}{c|}{MSRVTT} & \multicolumn{2}{c}{Avg} \\ 
        ~&&Acc & Score & Acc &  Score & Acc & Score & Acc & Score\\
        \midrule
        Video-LLaVA &183 & 47.0 & 3.34 & 69.7 &3.90 & 57.8 & 3.48 &58.1 & 3.57   \\ 
        \quad w/ PDrop &132 & 46.6 & 3.33 & 69.4 & 3.89 & 57.7 & 3.47 &57.9 & 3.56  \\

        \bottomrule
    \end{tabular}}
    \caption{GPT-Evaluation results on zero-shot video question answering Tasks. We apply PyramidDrop to accelerate the training process of vanilla Video-LLaVA model. The results show that we can achieve nearly a 30\% reduction in training time while maintaining comparable performance on video understanding tasks.}
    \label{tab:video_result}
\end{table}

\textbf{Ablation Studies}
In this part, we mainly study the influence of  $\lambda$ on both LLaVA-1.5 and LLaVA-NeXT.
Ablation studies about the number of stages S can be found in \textbf{Appendix}.
$\lambda$ balances the performance and efficiency of PyramidDrop, a larger $\lambda$ preserves more image information but slows down the training, and a smaller $\lambda$ has higher speedup while may influence the model performance.

As shown in Table~\ref{table:ablation}, we vary the $\lambda$ from 0.4 to 0.6 and report the model performance on both general and high-resolution benchmarks. 
For the general benchmarks, we observe a relative robust performance among different $\lambda$, this indicates that for most visual questions answering scenarios, our method is relatively robust to different hyperparameter choices, reducing the need for extensive trial and error to identify well-performing hyperparameter. When it comes to the DocVQA, which requires a fine-grained understanding on high-resolution images, the model performance shows a clear decline when the  $\lambda$ decreases to 0.4. It is reasonable due to the loss of critical image information and we could anticipate a more pronounced performance decline with the $\lambda$ keeps decreasing.
Therefore, we opt for \(\lambda = 0.5\), which maintains comparable performance while also yielding a significant reduction in processing time.

\section{Conclusion}

We introduce PyramidDrop, a simple yet effective strategy to reduce visual token redundancy in LVLMs, for boosting efficiency without performance loss. PyramidDrop helps to reduce the redundancy and concentrate more on valuable visual information for efficient deployment in realistic world. Our empirical study reveals that all visual tokens are necessary in the shallow layers of LVLMs, and token redundancy progressively increases in deeper layers. Experiments demonstrate that PyramidDrop can achieve up to 1.82$\times$ and 2.22$\times$ acceleration for training and inference respectively. 
{
    \small
    \bibliographystyle{ieeenat_fullname}
    \bibliography{main}
}

\newpage


\appendix
\section{Appendix}

\setlength{\tabcolsep}{1.5pt}
\begin{table*}[!t]
\renewcommand{\arraystretch}{1}
\small
\centering
\vspace{1pt}
\begin{tabular*}{0.85\textwidth}
{@{\extracolsep{\fill}}c|cccc|cccccc@{}}

 \toprule[1.2pt]
  Model & $\lambda$ & Stage & \makecell{GPU\\hours} & \makecell{Infer\\Flops(T)} & GQA & SEED$^I$ & MMB & \makecell{TextVQA}  & POPE & SQA \\ \midrule

\multirow{4}{*}{\makecell{LLaVA-1.5-7B}}      
 ~& vanilla & vanilla &104 (100\%)  & 3.82 & 62.0 & 66.1 & 64.3 & 58.2 & 85.9 & 66.8   \\ 
 ~& 0.5 & 3 & 85 (62.2\%) & 2.13 & \textbf{62.0} & \textbf{66.1} & \textbf{66.2} & 58.4 & \textbf{86.2} & 70.5  \\
 ~& 0.5 & 4 & 79 (76.0\%) & 1.78 & 61.9 & 65.5 & 66.1 & \textbf{58.5} & 86.0 & \textbf{71.0}  \\ 
 ~& 0.5 & 5 & 75 (78.9\%) & 1.38 & 61.4 & 65.5 & 65.9 & 57.8 & 86.1 & 69.9  \\ 

\bottomrule[1.2pt]
\end{tabular*}
\caption{\textbf{Ablation study results about stages S.} Dividing the LLM forward process into more stages causes the model to eliminate a larger number of image tokens in the earlier layers, leaving fewer tokens for processing in the later layers. On the other hand, using fewer stages reduces the number of token compression steps throughout the forward process, leading to increased redundancy. This parameter serves to balance the trade-off between the performance and efficiency of PyramidDrop.}
\vspace{-5pt}
\label{table:ablation stage}
\end{table*}

\section{Ablation Study about Stage S}
In this section, we primarily discuss the ablation study of stages \( S \). In these experiments, we set \(\lambda\) to 0.5, consistent with the previous experiments, and continue to follow the principle of evenly distributing layers within the LLM. If the entire LLM forward process is divided into more stages, the model will remove more image tokens at earlier layers, leaving fewer image tokens in the later layers of the LLM. Conversely, if fewer stages are used, the number of token compression steps during the forward process decreases, resulting in greater redundancy. This parameter is utilized to balance the performance and efficiency of PyramidDrop.

\subsection{Results Analysis}
As shown in Table~\ref{table:ablation stage}, we vary the number of stages from 3 to 5. Overall, the model's performance remains robust across these changes, demonstrating that our compression strategy is relatively well-designed and not overly sensitive to hyperparameters. 

However, on more challenging benchmarks such as SEED Bench and TextVQA, a noticeable performance decline occurs when the number of stages is increased to 5. If stages are further increased, the model's performance clearly deteriorates. This is reasonable because, at the maximum stage setting of 32, PyramidDrop would begin removing half of the image tokens right after the first layer, leaving only 2 image tokens by 8 layer, inevitably discarding critical image information. 

Meanwhile, with stages set to 3 or 4, there is no significant performance drop. Therefore, we ultimately select \( S = 4 \), which strikes a balance between preserving performance and effectively pruning redundancy by concentrating the limited image tokens on the important regions of the image."

\end{document}